%% file: main.tex
\documentclass[10pt,twocolumn,letterpaper]{article}

\usepackage{iccv}              
\usepackage{colortbl}
\usepackage{graphicx}
\usepackage{amsmath}
\usepackage{amssymb}
\usepackage{bm}
\usepackage{makecell}
\usepackage{multirow}  
\usepackage{array}     
\usepackage{wrapfig}
\usepackage{xcolor}    
\usepackage{adjustbox}
\usepackage{afterpage}
\usepackage{cuted}

\newcommand{\pub}[1]{{\color{gray}{\tiny{[{#1}]}}}}

\newcommand{\textgr}[1]{\textcolor[HTML]{006400}{\textbf{#1}}}

\usepackage{color}
\definecolor{iccvblue}{rgb}{0.21,0.49,0.74}
\definecolor{iccvred}{rgb}{0.8,0.2,0.2}
\definecolor{iccvgreen}{rgb}{0.2,0.8,0.2}
\definecolor{iccvpurple}{rgb}{0.5,0.2,0.8}
\definecolor{iccvorange}{rgb}{0.8,0.5,0.2}


\definecolor{iccvblue}{rgb}{0.21,0.49,0.74}
\usepackage[pagebackref,breaklinks,colorlinks,allcolors=iccvblue]{hyperref}

\title{DreamRenderer: Taming Multi-Instance Attribute Control in Large-Scale Text-to-Image Models}

\author{{Dewei Zhou$^{1}$, Mingwei Li$^{1}$, Zongxin Yang$^{2}$, Yi Yang$^{1*}$}\\
{RELER, CCAI, Zhejiang University $^{1}$ 
DBMI, HMS, Harvard University $^{2}$} \\
{\texttt{\{zdw1999,mingweili,yangyics\}@zju.edu.cn} \texttt{\{Zongxin Yang\}@hms.harvard.edu}} \\
{Project Page: \url{https://limuloo.github.io/DreamRenderer/}}
}
\newpage
\begin{document}
\maketitle
\input{Sections/0_abstract}    
\input{Sections/1_intro}

\input{Sections/2_related}
\input{Sections/3_method}
\input{Sections/4_experiments}
\input{Sections/5_conclusion}    
{
    \small
    \bibliographystyle{ieeenat_fullname}
    \bibliography{main}
}

\clearpage
\end{document}


\maketitle

\appendix
\input{Sections/6_appendix}
{
    \small
    \bibliographystyle{ieeenat_fullname}
    \bibliography{main}
}

%% file: Sections/0_abstract.tex
\begin{abstract}
Image-conditioned generation methods, such as depth- and canny-conditioned approaches, have demonstrated remarkable abilities for precise image synthesis. However, existing models still struggle to accurately control the content of multiple instances (or regions). Even state-of-the-art models like FLUX and 3DIS face challenges, such as attribute leakage between instances, which limits user control. To address these issues, we introduce DreamRenderer, a training-free approach built upon the FLUX model. DreamRenderer enables users to control the content of each instance via bounding boxes or masks, while ensuring overall visual harmony. We propose two key innovations: 1) Bridge Image Tokens for Hard Text Attribute Binding, which uses replicated image tokens as bridge tokens to ensure that T5 text embeddings, pre-trained solely on text data, bind the correct visual attributes for each instance during Joint Attention; 2) Hard Image Attribute Binding applied only to vital layers. Through our analysis of FLUX, we identify the critical layers responsible for instance attribute rendering and apply Hard Image Attribute Binding only in these layers, using soft binding in the others. This approach ensures precise control while preserving image quality. Evaluations on the COCO-POS and COCO-MIG benchmarks demonstrate that DreamRenderer improves the Image Success Ratio by 17.7\% over FLUX and enhances the performance of layout-to-image models like GLIGEN and 3DIS by up to 26.8\%. Project Page: https://limuloo.github.io/DreamRenderer/.
\end{abstract}

%% file: Sections/1_intro.tex
\section{Introduction}
\label{sec:intro}

Beyond conventional text-to-image models, image-conditioned generation methods~\cite{mou2024t2iad,controlnet}, such as depth- and canny-conditioned approaches, have emerged as powerful tools for more controllable content creation, finding applications in animation, game development, visual restoration, and virtual reality~\cite{controlnet,li2024controlnet++,xu2024ctrlora,zhang2024atlantis,stan2023ldm3d,liu2024smartcontrol,caphuman}.

However, existing techniques cannot accurately constrain the content generated within each specific instance (or region), and their performance deteriorates significantly as the number of controlled elements increases. As shown in Fig.~\ref{fig:teaser}, even state-of-the-art methods such as FLUX~\cite{flux} and 3DIS~\cite{zhou20243dis,zhou20253disflux} often fail to follow user-specified inputs, leading to unsatisfactory outcomes in which attributes bleed across different instances. Such limitations ultimately hinder more refined and controllable user-directed creation.

To address these challenges and grant users greater creative control, we introduce \textbf{DreamRenderer}, a training-free approach built upon the FLUX model~\cite{flux}. As illustrated in ~\cref{fig:teaser}, DreamRenderer functions as a \textbf{plug-and-play} tool, allowing users to further regulate the content of each instance via bounding boxes or masks, all while preserving overall visual harmony.

When implementing DreamRenderer, we encounter \textbf{two main challenges. 1) Ensuring text embeddings bind the correct visual information.}  Current state-of-the-art text-to-image models~\cite{flux,esser2024sd3} employ the T5 text encoder~\cite{T5}, pre-trained purely on textual data, to extract text embeddings lacking intrinsic visual information. These embeddings instead rely on Joint Attention~\cite{liu2024playground,dalva2024fluxspace,kong2024hunyuanvideo,yang2024cogvideox} with image embeddings to incorporate visual information~\cite{zhou20253disflux}. In scenarios where multiple instances are controlled, attribute confusion can easily arise among different instances and regions. \textbf{2) Rendering each instance accurately while preserving overall visual harmony.} To accurately generate controlled instances, it is necessary to constrain the image tokens’ attention masks~\cite{instancediffusion,rb,phung2024groundedrefocus} during Joint Attention. Yet overly restrictive constraints can undermine the resulting image quality, highlighting the need for more nuanced strategies to guide the attention masks of image tokens.

\input{Figures/02_Overview}

\textbf{A naive solution} to ensure each instance’s \textbf{text embedding binds the correct visual information} during Joint Attention is to constrain the tokens (both image and text) of a single instance to attend only to themselves. This strategy effectively simulates a single-instance generation process, thereby preserving the intended visual attributes for each text embedding. However, completely separating each instance's tokens from those of others significantly \textbf{degrades the image quality}, as shown in ~\cref{fig:ablation_text_binding}.

To address the two challenges outlined above, DreamRenderer introduces two key innovations: \textbf{1) Bridge tokens for Hard Text Attribute Binding.} Instead of directly using an instance’s original image tokens to help the text tokens bind the correct attributes, we replicate the instance’s image tokens, referred to as Bridge Image Tokens. These bridge tokens are not included in the model’s final output but instead simulate a single-instance generation process during Joint Attention, thereby guiding the text embeddings to acquire the correct visual attributes. In other words, each instance’s text embedding and Bridge Image Tokens attend only to each other, ignoring all other tokens. \textbf{2) Hard Image Binding only applied to vital layers.} After ensuring each text embedding accurately captures its instance’s visual attributes through Hard Text Attribute Binding, we further ensure each instance’s image embedding incorporates the correct visual information. We adopt two types of “image attribute binding”: a hard bind, where an instance’s image tokens attend exclusively to themselves, and a soft bind, where they can attend to all tokens in the image. From experimental observation, layers near the input and output of FLUX primarily encode global information, while the intermediate layers are crucial for rendering each instance. Therefore, we apply hard binding only in these middle layers, and soft binding elsewhere, thus preserving overall visual harmony while maintaining precise control over each instance’s attributes.

We evaluated DreamRenderer on two widely used benchmarks, COCO-POS~\cite{coco} and COCO-MIG~\cite{migc,migc++}, demonstrating its effectiveness in balancing image quality with fine-grained control. On COCO-POS, we extracted depth maps and canny maps from the COCO dataset as conditional inputs and leveraged their layouts to guide DreamRenderer. The results show that DreamRenderer boosts the Image Success Ratio by \textbf{17.7}\% over the original FLUX model, without sacrificing image quality. On COCO-MIG, we applied DreamRenderer to re-render the outputs of several state-of-the-art layout-to-image models, significantly enhancing both image quality and controllability. Specifically, DreamRenderer increases the Image Success Ratio for GLIGEN~\cite{gligen}, InstanceDiffusion~\cite{instancediffusion}, MIGC~\cite{migc}, and 3DIS~\cite{zhou20243dis} by \textbf{26.8}\%, \textbf{19.9}\%, \textbf{8.3}\%, and \textbf{7.4}\%, respectively.

The key contributions of this paper are as follows:
\begin{itemize}[leftmargin=*, noitemsep, topsep=0pt]

    \item We propose DreamRenderer, a training-free method that enables users to control the generation content for each region and instance in depth-conditioned or canny-conditioned generation.

    \item We introduce a novel Hard Text Attribute Binding technique, ensuring that text embeddings bind the correct visual attributes during the Joint Attention process.

    \item For multi-instance generation, we provide the first in-depth analysis of each layer’s potential function in the FLUX model, clarifying which layers handle global operations and which are pivotal for rendering individual instances, offering fresh insights for subsequent research.
\end{itemize}

%% file: Figures/02_Overview.tex
\begin{figure*}[tb!]
	\centering
	\includegraphics[width=1.0\linewidth]{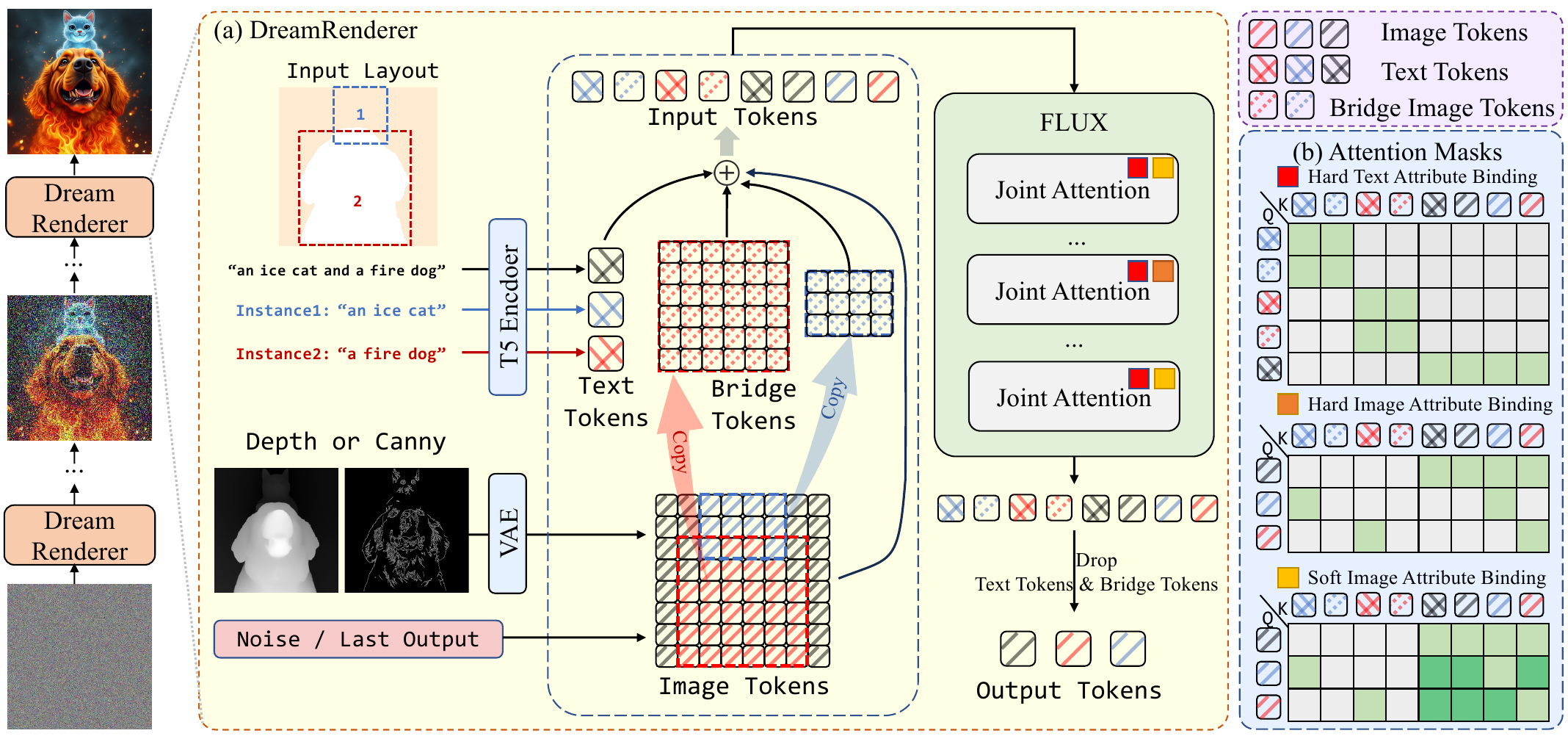}
	\vspace{-1.5mm}
	\caption{
		\textbf{The overview of DreamRenderer. (\S~\ref{sec:overview})} 
		\textbf{(a)} The pipeline of DreamRenderer. 
		\textbf{(b)} Attention maps in Joint Attention, which includes 1) Hard Text Attribute Binding \textbf{(\S~\ref{sec:hard_text_binding})}, 2) Hard Image Attribute Binding \textbf{(\S~\ref{sec:image_binding})}, and 3) Soft Image Attribute Binding \textbf{(\S~\ref{sec:image_binding})}.
		In the attention maps shown in (b), rows represent queries and columns represent keys.
		We use different \textbf{patterns} to distinguish between image tokens~\protect\adjustbox{valign=c}{\includegraphics[scale=0.4]{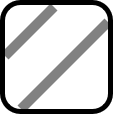}}~\protect\adjustbox{valign=c}{\includegraphics[scale=0.4]{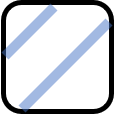}}~\protect\adjustbox{valign=c}{\includegraphics[scale=0.4]{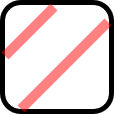}}, text tokens~\protect\adjustbox{valign=c}{\includegraphics[scale=0.4]{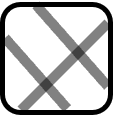}}~\protect\adjustbox{valign=c}{\includegraphics[scale=0.4]{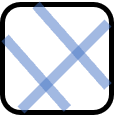}}~\protect\adjustbox{valign=c}{\includegraphics[scale=0.4]{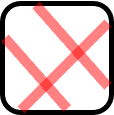}}, and bridge image tokens~\protect\adjustbox{valign=c}{\includegraphics[scale=0.4]{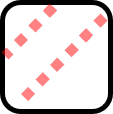}}~\protect\adjustbox{valign=c}{\includegraphics[scale=0.4]{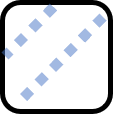}},
		while different \textbf{colors} (~\protect\adjustbox{valign=c}{\includegraphics[scale=0.4]{Sources/Text_02_Instance.png}}~\protect\adjustbox{valign=c}{\includegraphics[scale=0.4]{Sources/Image_02_Instance.png}}~\protect\adjustbox{valign=c}{\includegraphics[scale=0.4]{Sources/Bridge_02_Instance.png}} for an ice cat, ~\protect\adjustbox{valign=c}{\includegraphics[scale=0.4]{Sources/Text_03_Instance.png}}~\protect\adjustbox{valign=c}{\includegraphics[scale=0.4]{Sources/Image_03_Instance.png}}~\protect\adjustbox{valign=c}{\includegraphics[scale=0.4]{Sources/Bridge_01_Instance.png}} for a fire dog and~\protect\adjustbox{valign=c}{\includegraphics[scale=0.4]{Sources/Text_01_global.png}}~\protect\adjustbox{valign=c}{\includegraphics[scale=0.4]{Sources/Image_01_global.png}} for the global text tokens and background image tokens) represent tokens from different instances. 
	}
	\vspace{-1.5mm}
	\label{fig:overview}
\end{figure*}

%% file: Sections/2_related.tex
\section{Related Works}
\label{sec:related}

\noindent\textbf{Text-to-Image Generation.} In recent years, text-to-image generation has progressed rapidly~\cite{imagen,eDiff-I,DALL-E2,parti,glide,anydoor,lu2024mace,lu2023tf,gao2024eraseanything,Xie2024AddSRAD,li2024imagine,li2024anysynth,lu2024vine}, with generative models now producing high-quality images aligned with user descriptions. Early approaches—such as SD1.5~\cite{stablediffusion} and SDXL~\cite{podell2023sdxl}—adopted a U-Net-based~\cite{Unet,pydiff,zc_cycle,zhao2023wavelet,zhao2024learning} architecture together with a CLIP text encoder~\cite{CLIP} for extracting text embeddings, injecting textual information into the network via cross-attention~\cite{attention,structurediff} to ensure correspondence between text and output content. Later, the Diffusion Transformer (DiT)~\cite{diffedit} structure was introduced, replacing the U-Net with a Transformer, and using AdaLN to incorporate text information. To further refine text control, models like SD3~\cite{esser2024sd3} and FLUX~\cite{flux} switched to a T5-XXL~\cite{T5} text encoder, employing Joint Attention to align image and text embeddings and thereby ensure that the final outputs adhere closely to the textual prompts.

\noindent\textbf{Controllable Text-to-Image Generation.} With further improvements in image quality, users and researchers have turned their attention to more controllable content generation. Image-conditioned generation methods, like depth- and canny-conditioned techniques~\cite{controlnet,li2024controlnet++,bhat2024loosecontrol,lin2024ctrladapter,peng2024controlnext,wang2024omnicontrolnet,ip-adapter}, provide an outline-based or scene-level structure for generated images. However, current methods lack the ability to precisely regulate individual regions or instances—particularly in complex scenes with multiple controlled elements—often leading to inaccurate or undesirable results. To address this limitation, we introduce DreamRenderer, a plug-and-play controller that empowers users to specify the generation content of each region or instance with heightened accuracy. In addition, DreamRenderer can be applied to re-render the outputs of layout-to-image~\cite{gligen,instancediffusion,migc,zhou20243dis,wu2024ifadapter,zhang2024creatilayout,reco,ranni,ragdiff,lee2024groundit} models, further enhancing both content fidelity and overall image quality.

%% file: Sections/3_method.tex
\section{Method}
\subsection{Preliminaries}
\label{sec:pre}
\noindent\textbf{FLUX.} The FLUX~\cite{flux} model is a state-of-the-art text-to-image generation model that leverages the rectified flow~\cite{esser2024sd3,liu2022rflow} framework to iteratively generate images. In each iteration, FLUX first encodes the text prompt into a text embedding using a T5 text encoder~\cite{T5}, which is trained solely on textual data. This text embedding is then concatenated with a randomly sampled noise-based image embedding; if needed, a control image embedding (e.g., depth or canny) is also included. Subsequently, multiple layers of \textbf{Joint Attention}~\citep{liu2024playground,dalva2024fluxspace,zhou20253disflux} iteratively refine both the image and text embeddings, ensuring they remain in alignment. Finally, FLUX decodes the refined image embedding through a Variational Autoencoder (VAE)~\cite{vae}, generating an output image that faithfully reflects the user’s textual description.

\noindent\textbf{Joint Attention.} In Joint Attention, both text and image embeddings are processed simultaneously. Each embedding is first transformed into token representations via a linear layer: the text embedding is converted into $Q_{\text{text}}$, $K_{\text{text}}$, and $V_{\text{text}}$, while the image embedding is mapped to $Q_{\text{image}}$, $K_{\text{image}}$, and $V_{\text{image}}$. These token sets are then concatenated as follows: $Q = Q_{\text{text}} \oplus Q_{\text{image}}$, $K = K_{\text{text}} \oplus K_{\text{image}}$, and $V = V_{\text{text}} \oplus V_{\text{image}}$, where $\oplus$ denotes the concatenation operation. The attention mechanism is subsequently applied to the concatenated tokens.
\vspace{-2mm}
\begin{equation}
	\text{Attention}(Q, K, V) = \text{Softmax}\left(\frac{QK^T}{\sqrt{d_k}}\right)V
\end{equation}
where $d_k$ is the dimension of the key vectors. This Joint attention mechanism allows for bidirectional information flow between text and image embeddings, enabling the model to align visual content with textual descriptions.

\subsection{Problem Definition and Challenges}

\noindent{\textbf{Problem Definition.}} 
Given a depth/canny map, a global prompt, and detailed instance (or region) descriptions, the objective is to generate an image that not only adheres to the global structural constraints of the depth/canny map and global prompt, but also ensures that each controlled instance accurately adheres the specified descriptions.

\noindent{\textbf{Challenges.}} We build DreamRenderer upon the FLUX model (\textbf{\S~\ref{sec:pre}}), and there are two main challenges in achieving our objectives: \textit{1) Correctly binding text embedding attributes in multi-instance scenarios.}
Unlike previous text-to-image models that use CLIP~\cite{CLIP}, FLUX employs the T5 text encoder, which is pre-trained on pure text data and does not contain visual information. This creates a challenge in ensuring that the text embedding correctly binds the visual attributes of each instance during the Joint Attention process, particularly when multiple instances are involved. \textit{2) Enhancing the overall harmony of the generated image while maintaining correct attributes for each instance.} In multi-instance scenarios, the image tokens corresponding to different instances may inadvertently leak attributes~\cite{phung2024groundedrefocus} during Joint Attention. Our goal is to minimize such leakage while ensuring that the final generated image retains a cohesive and harmonious composition across all instances.

\input{Figures/05_Ablation}

\subsection{Overview}
\label{sec:overview}

Fig.~\ref{fig:overview} illustrates the overview of DreamRenderer. In the Joint Attention mechanism, DreamRenderer introduces a novel $\textbf{Hard Text Attribute Binding (\S\ref{sec:hard_text_binding})}$ algorithm to ensure that the text embedding for each instance correctly binds the relevant visual information. Additionally, to enhance the overall harmony of the generated image while maintaining accurate image embedding attributes for each instance, we conducted an $\textbf{experimental analysis (\S\ref{sec:image_binding})}$ of each layer in FLUX and decided to apply $\textbf{Hard Image Attribute Binding (\S\ref{sec:image_binding})}$ only in the middle layers of the FLUX model. In all other layers, $\textbf{Soft Image Attribute Binding (\S\ref{sec:image_binding})}$ is used.

\subsection{Preparation}
\label{sec:preparation}

As shown in Fig.~\ref{fig:overview} (a), DreamRenderer first embeds the input text descriptions for each instance and the global prompt \textbf{separately} through the T5 text encoder. These encoded embeddings are then \textbf{concatenated} to form the complete text embedding for the generation process. Our method requires users to provide either a depth map or a canny edge map as structural guidance, which serves as the foundation for the spatial arrangement of instances in the generated image. For instance localization, we utilize bounding boxes or masks provided by the user to identify each instance's region within this structural guidance.

\subsection{Hard Text Attribute Binding}
\label{sec:hard_text_binding}

\noindent\textbf{Motivation.} When generating a single instance, the FLUX model generally produces images that align with the textual prompt, exhibiting minimal attribute errors. In such scenarios, image and text tokens in Joint Attention focus exclusively on the information of that single instance, thereby enabling the text embedding to bind accurate visual attributes. Building on this insight, we propose that \textit{in multi-instance scenarios, the image and text tokens of each instance should primarily attend to themselves rather than to tokens belonging to different instances, allowing the text embedding to effectively bind the correct visual information}.

\noindent\textbf{Naive Solution.} A straightforward approach to ensure that each instance’s text embedding is bound to the correct attributes is to process each instance independently during Joint Attention. In this method, both image and text tokens for a given instance interact exclusively with themselves, remaining isolated from tokens of other instances. However, this complete isolation introduces a significant \textbf{drawback}: it disrupts the visual harmony of the overall image and substantially lowers the quality of the generated result (as shown in Fig.~\ref{fig:ablation_text_binding}).

\noindent\textbf{Bridge Image Token for Advanced Solution.} Since strictly isolating original image tokens for each instance in joint Attention degrades the image quality, DreamRenderer proposes an advanced solution: during Joint Attention, an additional copy of each instance's image tokens are created, referred to as Bridge Image Tokens. These Bridge Image Tokens do not contribute to the final output image but serves solely to assist each instance's text embedding in binding the correct visual attributes during the Joint Attention. As shown in \cref{fig:overview}, the Bridge Image Tokens and text tokens for each instance align exactly as in a single-instance generation process, ensuring that the visual attributes in the final text embedding are consistent with the textual description. 
Formally, for the $i$-th instance, the Hard Text Attribute Binding attention mask $M_{\text{text}}^i$ is defined as:
\begin{equation}
    M_{\text{text}}^i[q,k] = \begin{cases}
        1, & \text{if } q,k \in \mathcal{T}_i \cup \mathcal{B}_i \\
        0, & \text{otherwise}
    \end{cases}
\end{equation}
where $q$ and $k$ are the query and key indices respectively, $\mathcal{T}_i$ and $\mathcal{B}_i$ denote the token indices of the $i$-th instance's text and Bridge Image Tokens respectively.
\subsection{Image Attribute Binding}
\label{sec:image_binding}

\noindent\textbf{Overview.} After ensuring the accuracy of the text embedding's attributes, the next step is to guarantee the correctness of the visual attributes in each instance's image tokens. As shown in ~\cref{fig:overview}, DreamRenderer employs Hard Image Attribute Binding at the vital binding layer to ensure that each instance is rendered with the correct attributes. In the remaining layers, Soft Image Attribute Binding is used to ensure that all instances ultimately form a coherent image. In the following sections, we will detail the mechanisms of hard and soft image attribute binding and explain how we identify the vital layers for Hard Image Attribute Binding.

\noindent\textbf{Hard Image Attribute Binding.} As shown in~\cref{fig:overview}, the Hard Image Attribute Binding imposes attention mask constraints on each instance’s image tokens during Joint Attention. First, it ensures that each instance attends only to its corresponding text token—which, following Hard Text Attribute Binding, contains the correct visual information. Second, to prevent attribute leakage across different image tokens, Hard Image Attribute Binding further restricts each instance’s image tokens to attend its own image tokens.
For the $i$-th instance, the Hard Image Attribute Binding mask $M_{\text{hard}}^i$ is constructed as:
\begin{equation}
    M_{\text{hard}}^i[q,k] = \begin{cases}
        1, & \text{if } q \in \mathcal{I}_i, k \in \mathcal{T}_i \cup \mathcal{I}_i \\
        0, & \text{otherwise}
    \end{cases}
\end{equation}
where $\mathcal{I}_i$ represents the set of image token indices belonging to the $i$-th instance, and $\mathcal{T}_i$ represents its corresponding text token indices.

\noindent\textbf{Soft Image Attribute Binding.} To ensure the overall coherence of the final generated image, Soft Image Attribute Binding relaxes the constraints imposed by Hard Image Attribute Binding. Specifically, for each instance's image tokens, Soft Image Attribute Binding allows it to attend to the image tokens of the entire image, thereby enhancing the cohesion of the entire scene.
For the $i$-th instance, the Soft Image Attribute Binding mask $M_{\text{soft}}^i$ is constructed as:
\begin{equation}
    M_{\text{soft}}^i[q,k] = \begin{cases}
        1, & \text{if } q \in \mathcal{I}_i , k \in \mathcal{I}_{\text{all}} \\
        0, & \text{otherwise}
    \end{cases}
\end{equation}
where $\mathcal{I}_{\text{all}} = \mathcal{I}_\text{background} \cup \left(\bigcup_{i=1}^N \mathcal{I}_i\right)$ represents the set of all image token indices, $\mathcal{I}_\text{background}$ is the set of all image tokens that are not associated with any instance.

\input{Figures/07_CocoPos}

\noindent\textbf{Search Vital Binding Layers.} As shown in ~\cref{fig:search_vital}, we applied Hard Image Attribute Binding layer by layer on the FLUX network, which consists of 57 Joint Attention layers, and compared the result obtained using Soft Image Binding across all layers to identify which layers are more suitable for binding specific instance attributes. The results from Fig.~\ref{fig:search_vital} indicate that applying Hard Image Binding near the output and input layers of FLUX leads to a significant decrease in performance. 
Conversely, we observed that implementing hard image attribute binding in the middle layers of FLUX often enhances attribute fidelity. Based on these findings, we argue that the layers at the input and output of FLUX are primarily involved in handling global image information, while the mid-layers play a critical role in rendering the attributes of instances within the image. Therefore, we perform Hard Image Binding in the mid-layers of FLUX, while employing Soft Image Binding in the remaining layers. This approach optimally balances the fidelity of instance attributes with the overall coherence of the image.

%% file: Figures/05_Ablation.tex
\begin{figure}[t!]
	\centering
	\includegraphics[width=1.0\linewidth]{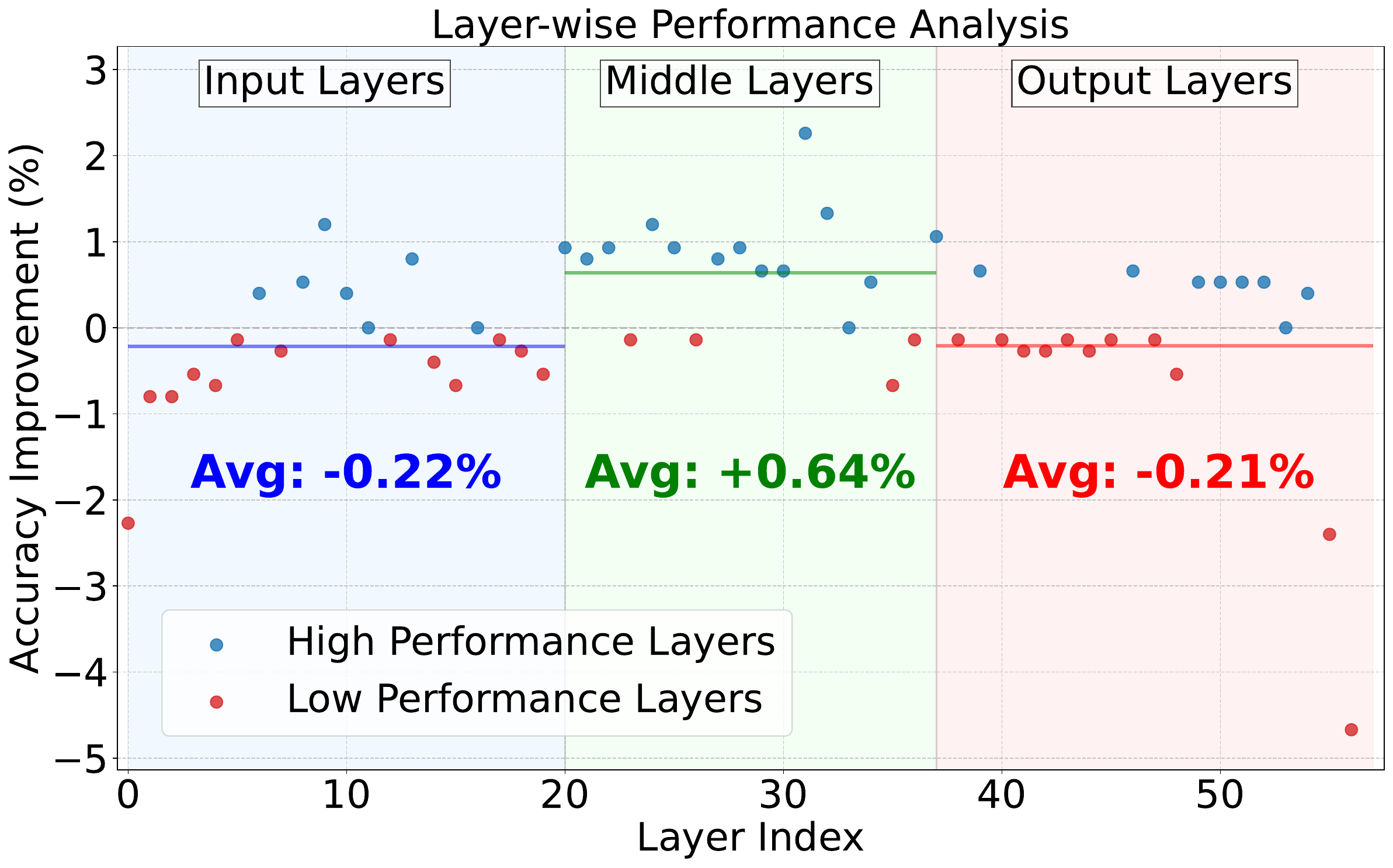}
	\vspace{-6mm}
	\caption{
        \textbf{Vital Binding Layer Search (\S~\ref{sec:image_binding}).} 
        We apply Hard Image Attribute Binding layer by layer and observe that applying it in the FLUX model’s input or output layers degrades performance, whereas applying it in the middle layers yields improvements.
	}
    \vspace{-5mm}
	\label{fig:search_vital}
\end{figure}

%% file: Figures/07_CocoPos.tex
\begin{figure*}[tb!]
	\centering
	\includegraphics[width=1.0\linewidth]{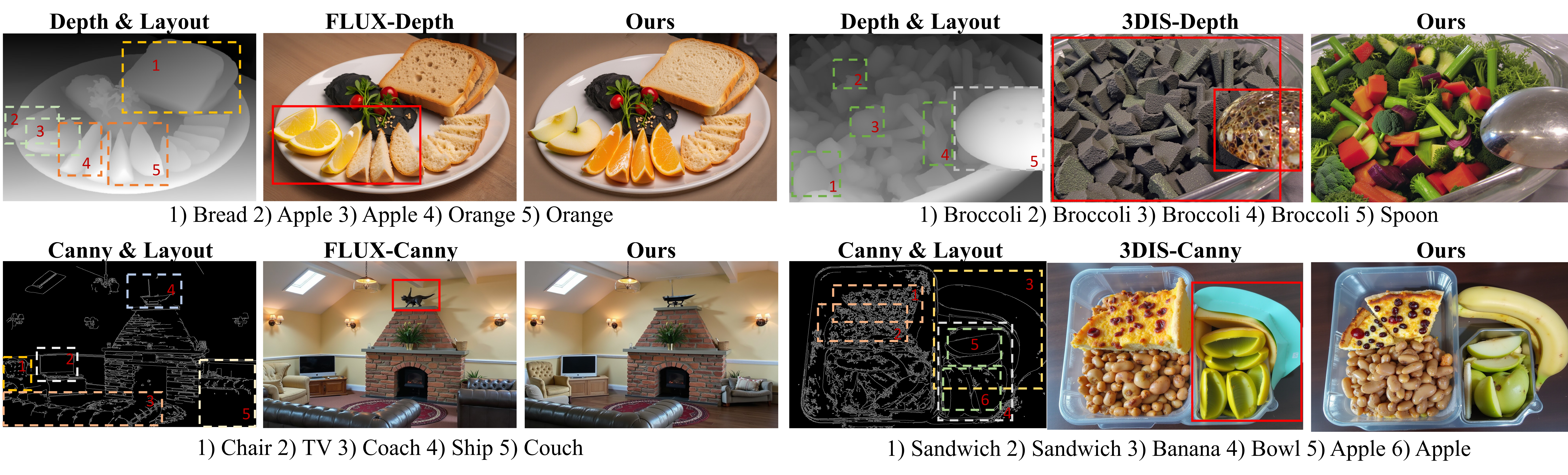}
	\vspace{-8mm}
	\caption{
		\textbf{Qualitative Comparison on the COCO-POS benchmark (\S~\ref{sec:comparison}).}
	}
	\vspace{-1.5mm}
	\label{fig:coco_pos}
\end{figure*}

%% file: Sections/4_experiments.tex
\section{Experiments}
\label{sec:experiments}
\subsection{Experiment Setup}
\label{sec:experiment_setup}

\input{Tables/main_results}

\input{Tables/table_1}
\input{Figures/MIG_VIS}
\textbf{Baselines.}
Beside the FLUX model, we evaluate our method against several state-of-the-art methods for multi-instance generation. Since DreamRenderer is designed as a \textbf{plug-and-play} solution, we conduct experiments by integrating it with existing methods: GLIGEN~\cite{gligen}, InstanceDiffusion~\cite{instancediffusion}, MIGC~\cite{migc}, and 3DIS~\cite{zhou20253disflux}.

\noindent\textbf{Implementation Details.}
We perform canny-conditioned and depth-conditioned generation using FLUX.1-Canny~\cite{flux} and FLUX.1-Depth~\cite{flux}, respectively. For both cases, we sample images over 20 steps. During depth-conditioned generation, we set the Classifier-Free Guidance (CFG)~\cite{Ho2022ClassifierFreeDG} scale to 10.0, while for canny-conditioned generation, the CFG scale is set to 30. In our experiments, for instances whose positions are specified by bounding boxes, we further segment them using the SAM-2~\cite{sam,ravi2024sam2} model to obtain more precise instance masks, which is consistent with previous work~\cite{zhou20243dis}.

\noindent\textbf{Evaluation Benchmarks.}
We conduct experiments on two widely used benchmarks: \textbf{1) COCO-POS Benchmark}, which requires generating images according to specified layouts. We extract depth maps or canny edges from COCO dataset images as conditioning signals and utilize the dataset's inherent layouts for rendering. Models must generate results matching the instance categories at designated locations. Here we compare our approach with training-free rendering methods including Multi-Diffusion and 3DIS. \textbf{2) COCO-MIG Benchmark}, which tests multi-instance generation with accurate position and attribute control. We evaluate DreamRenderer's integration capabilities with state-of-the-art MIG models by first generating RGB images with these models, then extracting depth maps to combine with layouts for instance rendering. This evaluates DreamRenderer's attribute control effectiveness when applied to existing MIG frameworks.

\noindent\textbf{Evaluation Metrics.} We used the following metrics to evaluate the model: \textit{1) Mean Intersection over Union (MIoU)}: Measures the overlap ratio between the rendered instance positions and the target positions. \textit{2) Local CLIP Score}: Assesses the visual consistency of the rendered instances with their corresponding textual descriptions. \textit{3) Average Precision (AP)}: Evaluates the accuracy of the rendered image's layout.  \textit{4) Instance Success Ratio (ISR)}: Calculates the ratio of correctly rendered instances. \textit{5) Image Success Ratio (ISR)}: Measures the ratio of images where all instances are correctly rendered.
\subsection{Comparison with State-of-the-Art Methods}
\label{sec:comparison}
\noindent\textbf{COCO-POS Benchmarks.} 
Tab.~\ref{tab:coco_pos} presents quantitative results comparing our method against FLUX and 3DIS. Our approach achieves consistent superiority across all metrics in both depth-guided and canny-guided generation scenarios. In the depth-guided setting, DreamRenderer shows a substantial improvement in SR (\textbf{62.50\%} vs. \textbf{53.88\%} for 3DIS), indicating more coherent scene structures. The high ISR (\textbf{94.51\%}) and MIoU (\textbf{84.36\%}) further corroborate its precise instance-level control. In the more demanding canny-guided scenario, DreamRenderer also exceeds the 3DIS by \textbf{5.21\%} in SR. Meanwhile, as shown in Fig.~\ref{fig:coco_pos}, our method does not compromise the original FLUX model’s image generation quality—thanks to applying hard image attribute binding only within the vital layers.

\noindent\textbf{COCO-MIG Benchmarks.} Tab.~\ref{tab:coco_mig} and Fig.~\ref{fig:mig_vis} present the results of applying DreamRenderer to various state-of-the-art layout-to-image methods. As shown, DreamRenderer substantially enhances the instance-attribute control accuracy, ultimately boosting the Image Success Ratio by \textbf{26.8\%} over GLIGEN, \textbf{19.9\%} over InstanceDiffusion, \textbf{8.3\%} over MIGC, and \textbf{7.4\%} over 3DIS. Notably, this improvement becomes more pronounced as the number of instances to be controlled increases: for example, DreamRenderer's performance gain over 3DIS is only \textbf{2.5\%} when controlling two instances, but rises to \textbf{10.5\%} when controlling six. These benefits stem from our Hard Text Attribute Binding algorithm, which ensures that each instance’s text embedding accurately binds its visual attributes during Joint Attention, even for a large number of instances.

\input{Tables/UserStudy}
\input{Tables/Ablation_Text_Binding}
\input{Tables/Ablation_Image_Binding}

\input{Figures/04_Ablation_Image_Binding}
\input{Figures/04_Ablation_Text_Binding}
\noindent\textbf{User Study.} Tab.~\ref{tab:user_study} shows a user study with \textbf{31} participants comparing our method to FLUX~\cite{flux} and 3DIS~\cite{zhou20253disflux} on perceptual quality. Participants viewed paired outputs and rated each on (1) \textit{Layout Accuracy} and (2) \textit{Image Quality}, using a 5-point scale, in a blind comparison with randomized presentation. Each participant evaluated 17 pairs, with the input layout and text descriptions displayed. The results show that our proposed DreamRenderer not only enhances the FLUX model’s layout control capabilities but also generates outputs that are more visually appealing to users.

\subsection{Ablation Study}
\label{sec:ablation}

\textbf{Bridge Image Tokens for Hard Text Attribute Binding.}
Tab.~\ref{tab:ablation_text_binding} and Fig.~\ref{fig:ablation_text_binding} present our ablation study on the Hard Text Attribute Binding mechanism. The Naive Solution (\S~\ref{sec:hard_text_binding}) isolates each instance during Joint Attention, disrupting the model’s inherent feature distribution and thus causing a performance drop. Introducing Bridge Image Tokens—which are not part of the final output—can effectively address this issue, enabling text tokens to bind the correct attributes and improving accuracy without compromising image quality. As the number of controlled instances increases, the benefit of Hard Text Attribute Binding becomes more pronounced: for example, moving from 2 to 6 instances raises the Instance Success Ratio improvement from \textbf{3.5}\% to \textbf{6.2}\%.

\noindent\textbf{Vital Layers for Image Attribute Binding.}
\cref{tab:ablation_image_binding} and \cref{fig:ablation_image_binding} present our ablation study on the Hard Image Attribute Binding mechanism. Applying Hard Image Attribute Binding at the FLUX input or output layers yields no clear performance gains and substantially degrades image quality, indicating that these layers are critical for the model's global information processing. Imposing instance or region isolation at these stages severely disrupts the intermediate feature distribution, ultimately causing a sharp drop in performance. In contrast, restricting Hard Image Attribute Binding to the mid layers preserves image quality while significantly improving performance—for instance, boosting the Instance Success Ratio by \textbf{15.7}\%. This finding shows that FLUX's mid layers play a pivotal role in determining each instance’s visual content, making them more suitable for binding instance's attribute.

%% file: Tables/main_results.tex
\begin{table}[tb!]
	\centering
	\caption{\textbf{Quantitative results on COCO-POS benchmark (\S~\ref{sec:comparison})}. SR: Success Ratio of the Entire Image, ISR: Instance Success Ratio, MIoU: Mean Intersection over Union, AP: Average Precision. $*$: canny-guided. $\dagger$: depth-guided.}\label{tab:coco_pos}
	\vspace{-3.5mm}
	\centering
	\resizebox{\columnwidth}{!}{
		\setlength{\tabcolsep}{5pt}
		\footnotesize
		\begin{tabular}{rccccc}
			\bottomrule[1pt]
            \rowcolor[HTML]{FAFAFA}
			Method & SR (\%)$\uparrow$ & ISR (\%)$\uparrow$ & MIoU (\%)$\uparrow$ & AP (\%)$\uparrow$ & CLIP$\uparrow$ \\ \toprule[0.8pt]
			\rowcolor[HTML]{F8FBFF} FLUX$^*$\hspace{0.3em}\hspace{0.1em} & 16.38 & 64.95 & 59.71 & 31.88 & 19.03 \\
			\rowcolor[HTML]{F8FBFF} 3DIS$^*$\hspace{0.30em}\hspace{0.1em} & 18.07 & 68.78 & 62.97 & 31.36 & 19.46 \\
			\rowcolor[HTML]{F8FBFF} Ours$^*$\hspace{0.30em}\hspace{0.1em} & \textbf{23.28} & \textbf{74.61} & \textbf{66.95} & \textbf{37.00} & \textbf{20.03} \\
			\hline
			\rowcolor[HTML]{F8FFF8} FLUX$^\dagger$\hspace{0.30em}\hspace{0.1em} & 44.83 & 85.13 & 76.86 & 50.72 & 19.82 \\
			\rowcolor[HTML]{F8FFF8} 3DIS$^\dagger$\hspace{0.30em}\hspace{0.1em} & 53.88 & 90.33 & 81.26 & 54.26 & 20.23 \\
			\rowcolor[HTML]{F8FFF8} Ours$^\dagger$\hspace{0.30em}\hspace{0.1em} & \textbf{62.50} & \textbf{94.51} & \textbf{84.36} & \textbf{58.95} & \textbf{20.74} \\
			\toprule[0.8pt]
		\end{tabular}
	}
	\vspace{-4.0mm}
\end{table}

%% file: Tables/table_1.tex
\begin{table*}[tb!]
	\centering
	\caption{\textbf{Quantitative results on COCO-MIG benchmark (\S~\ref{sec:comparison})}. $\mathcal{L}_i$ means that the count of instances needed to generate is \textbf{i}.}\label{tab:coco_mig}
	\vspace{-3.5mm}
	\centering
	\resizebox{\textwidth}{!}{
	\setlength{\tabcolsep}{3.8pt}
	\footnotesize
	\renewcommand\arraystretch{1.1}
	\begin{tabular}{ccccccccccccccccccc}
		\bottomrule[1pt]\rowcolor[HTML]{FAFAFA}
		                                                                              & \multicolumn{6}{c}{Instance Success Ratio (\%)$\uparrow$} & \multicolumn{6}{c}{Mean Intersection over Union (\%)$\uparrow$} & \multicolumn{6}{c}{Image Success Ratio (\%)$\uparrow$}                                                                                                                                                                             \\
		\cmidrule(lr){1-1} \cmidrule(lr){2-7} \cmidrule(lr){8-13} \cmidrule(lr){14-19}
		Method & $\mathcal{L}_2$ & $\mathcal{L}_3$ & $\mathcal{L}_4$ & $\mathcal{L}_5$ & $\mathcal{L}_6$ & $\mathcal{AVG}$ & $\mathcal{L}_2$ & $\mathcal{L}_3$ & $\mathcal{L}_4$ & $\mathcal{L}_5$ & $\mathcal{L}_6$ & $\mathcal{AVG}$ & $\mathcal{L}_2$ & $\mathcal{L}_3$ & $\mathcal{L}_4$ & $\mathcal{L}_5$ & $\mathcal{L}_6$ & $\mathcal{AVG}$ \\ \toprule[0.8pt]

		\rowcolor[HTML]{F8FFF8} GLIGEN\hspace{0.30em}~\pub{CVPR23}\hspace{0.1em} & 41.6 & 31.6 & 27.0 & 28.3 & 27.7 & 29.7 & 37.1 & 29.2 & 25.1 & 26.7 & 25.5 & 27.5 & 16.8 & 4.6 & 0.0 & 0.0 & 0.0 & 4.4 \\
		\rowcolor[HTML]{F8FFF8} +Ours & 75.5 & 73.6 & 67.9 & 62.6 & 66.2 & 67.8 & 64.3 & 63.5 & 58.1 & 54.2 & 56.6 & 58.2 & 57.4 & 45.1 & 27.0 & 11.4 & 13.0 & 31.2 \\
		\rowcolor[HTML]{F8FFF8} \multicolumn{1}{c}{vs. GLIGEN} & \textgr{+33.9} & \textgr{+42.0} & \textgr{+40.9} & \textgr{+34.3} & \textgr{+38.5} & \textgr{+38.1} & \textgr{+27.2} & \textgr{+34.3} & \textgr{+33.0} & \textgr{+27.5} & \textgr{+31.1} & \textgr{+30.7} & \textgr{+40.6} & \textgr{+40.5} & \textgr{+27.0} & \textgr{+11.4} & \textgr{+13.0} & \textgr{+26.8} \\

		\hline

		\rowcolor[HTML]{F8FBFF} InstanceDiff\hspace{0.3em}~\pub{CVPR24}\hspace{0.1em} & 58.4 & 51.9 & 55.1 & 49.1 & 47.7 & 51.3 & 53.0 & 47.4 & 49.2 & 43.7 & 42.6 & 46.0 & 37.4 & 15.0 & 11.5 & 5.0 & 2.6 & 14.5 \\
		\rowcolor[HTML]{F8FBFF} +Ours & 77.4 & 75.4 & 72.5 & 65.3 & 67.2 & 70.1 & 68.8 & 66.4 & 63.0 & 57.8 & 57.5 & 61.2 & 58.7 & 45.8 & 34.5 & 13.6 & 17.5 & 34.4 \\
		\rowcolor[HTML]{F8FBFF} \multicolumn{1}{c}{vs. InstanceDiff} & \textgr{+19.0} & \textgr{+23.5} & \textgr{+17.4} & \textgr{+16.2} & \textgr{+19.5} & \textgr{+18.8} & \textgr{+15.8} & \textgr{+19.0} & \textgr{+13.8} & \textgr{+14.1} & \textgr{+14.9} & \textgr{+15.2} & \textgr{+21.3} & \textgr{+30.8} & \textgr{+23.0} & \textgr{+8.6} & \textgr{+14.9} & \textgr{+19.9} \\

		\hline

		\rowcolor[HTML]{FFFCF5} MIGC\hspace{0.30em}~\pub{CVPR24}\hspace{0.1em} & 74.8 & 66.2 & 67.2 & 65.3 & 66.1 & 67.1 & 64.5 & 56.9 & 57.0 & 55.5 & 57.3 & 57.5 & 54.8 & 32.7 & 22.3 & 12.9 & 17.5 & 28.4 \\
		\rowcolor[HTML]{FFFCF5} +Ours & 79.7 & 76.3 & 70.6 & 67.7 & 69.4 & 71.4 & 68.8 & 65.0 & 60.3 & 57.6 & 59.3 & 61.0 & 64.5 & 47.7 & 30.4 & 18.6 & 20.1 & 36.7 \\
		\rowcolor[HTML]{FFFCF5} \multicolumn{1}{c}{vs. MIGC} & \textgr{+4.9} & \textgr{+10.1} & \textgr{+3.4} & \textgr{+2.4} & \textgr{+3.3} & \textgr{+4.3} & \textgr{+4.3} & \textgr{+8.1} & \textgr{+3.3} & \textgr{+2.1} & \textgr{+2.0} & \textgr{+3.5} & \textgr{+9.7} & \textgr{+15.0} & \textgr{+8.1} & \textgr{+5.7} & \textgr{+2.6} & \textgr{+8.3} \\

		\hline

		\rowcolor[HTML]{FFF8F0} 3DIS\hspace{0.30em}~\pub{ICLR25}\hspace{0.1em} & 76.5 & 68.4 & 63.3 & 58.1 & 58.9 & 62.9 & 67.3 & 61.2 & 56.4 & 52.3 & 52.7 & 56.2 & 61.3 & 36.0 & 25.0 & 12.9 & 11.7 & 29.7 \\
		\rowcolor[HTML]{FFF8F0} +Ours & 79.0 & 77.3 & 71.8 & 68.7 & 69.4 & 71.9 & 69.8 & 68.2 & 63.7 & 61.4 & 61.2 & 63.7 & 63.2 & 52.9 & 29.7 & 18.6 & 18.8 & 37.1 \\
		\rowcolor[HTML]{FFF8F0} \multicolumn{1}{c}{vs. 3DIS} & \textgr{+2.5} & \textgr{+8.9} & \textgr{+8.5} & \textgr{+10.6} & \textgr{+10.5} & \textgr{+9.0} & \textgr{+2.5} & \textgr{+7.0} & \textgr{+7.3} & \textgr{+9.1} & \textgr{+8.5} & \textgr{+7.5} & \textgr{+1.9} & \textgr{+16.9} & \textgr{+4.7} & \textgr{+5.7} & \textgr{+7.1} & \textgr{+7.4} \\

		\toprule[0.8pt]
	\end{tabular}
	}
	\vspace{-2mm}
\end{table*}

%% file: Figures/MIG_VIS.tex
\begin{figure*}[ht!]
	\centering
	\includegraphics[width=1.0\linewidth]{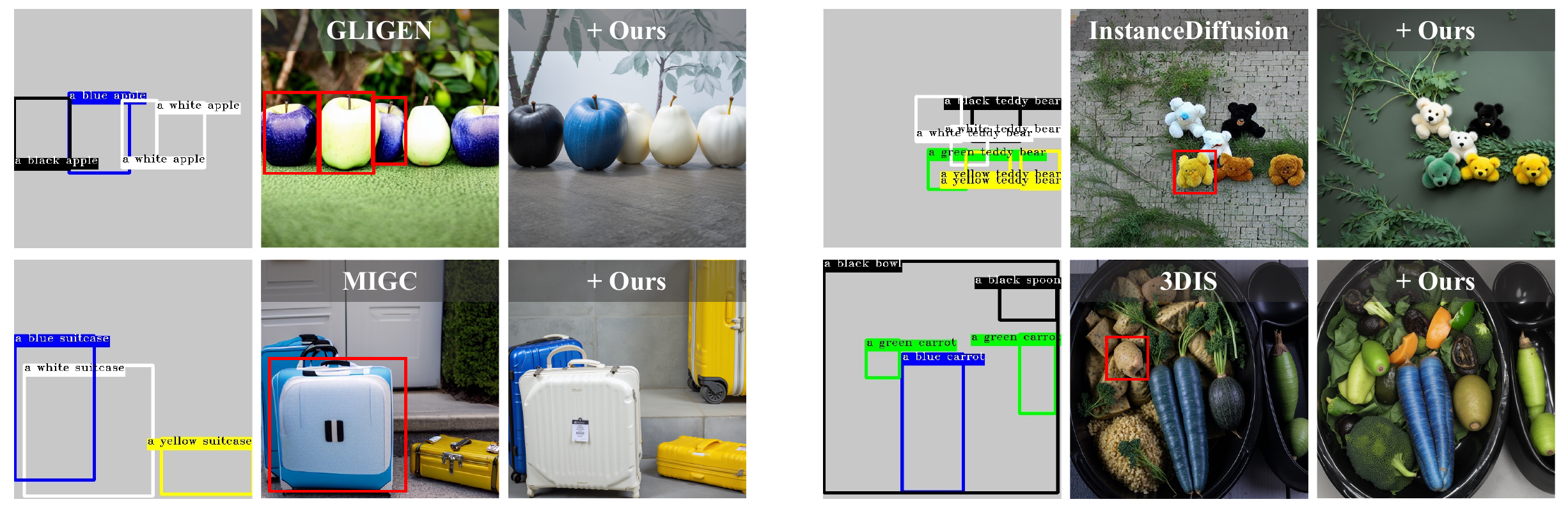}
    \vspace{-5mm}
	\caption{\textbf{Qualitative comparison on the COCO-MIG benchmark (\S~\ref{sec:comparison}).}
	}
	\label{fig:mig_vis}
\end{figure*}

%% file: Tables/UserStudy.tex
\begin{table}[ht!]
	\centering
	\caption{\textbf{User study results} with 31 participants (1-5 scale)(\S~\ref{sec:comparison}).}\label{tab:user_study}
	\vspace{-4mm}
	\centering
	\resizebox{0.8\columnwidth}{!}{
		\setlength{\tabcolsep}{3pt}
		\scriptsize
		\begin{tabular}{rccc}
			\bottomrule[1pt]
            \rowcolor[HTML]{FAFAFA}
			Method & Layout Accuracy$\uparrow$ & Visual Quality$\uparrow$ \\ \toprule[0.8pt]
			\rowcolor[HTML]{F8FBFF} FLUX & 3.51 & 3.26 \\
			\rowcolor[HTML]{F8FBFF} 3DIS & 3.63 & 3.31 \\
			\rowcolor[HTML]{FFF8F0} Ours & \textbf{4.04} & \textbf{3.80} \\
			\toprule[0.8pt]
		\end{tabular}
	}
	\vspace{-6pt}
\end{table}

%% file: Tables/Ablation_Text_Binding.tex
\begin{table*}[t!]
	\centering
	\caption{\textbf{Ablation study on Hard Text Attribute Binding (\S~\ref{sec:hard_text_binding})}. $\mathcal{L}_i$ means that the count of instances needed to generate is \textbf{i}.}
	\label{tab:ablation_text_binding}
	\vspace{-3.5mm}
	\centering
	\resizebox{\textwidth}{!}{
		\setlength{\tabcolsep}{3.8pt}
		\footnotesize
		\renewcommand\arraystretch{1.1}
		\begin{tabular}{ccccccccccccccccccc}
			\bottomrule[1pt]\rowcolor[HTML]{FAFAFA}
			                                & \multicolumn{6}{c}{Instance Success Ratio (\%)$\uparrow$} & \multicolumn{6}{c}{Mean Intersection over Union (\%)$\uparrow$} & \multicolumn{6}{c}{Image Success Ratio (\%)$\uparrow$}                                                                                                                                                                   \\
			\cmidrule(lr){1-1} \cmidrule(lr){2-7} \cmidrule(lr){8-13} \cmidrule(lr){14-19}
			Method                          & $\mathcal{L}_2$ & $\mathcal{L}_3$ & $\mathcal{L}_4$ & $\mathcal{L}_5$ & $\mathcal{L}_6$ & $\mathcal{AVG}$ & $\mathcal{L}_2$ & $\mathcal{L}_3$ & $\mathcal{L}_4$ & $\mathcal{L}_5$ & $\mathcal{L}_6$ & $\mathcal{AVG}$ & $\mathcal{L}_2$ & $\mathcal{L}_3$ & $\mathcal{L}_4$ & $\mathcal{L}_5$ & $\mathcal{L}_6$ & $\mathcal{AVG}$ \\ \toprule[0.8pt]

			\rowcolor[HTML]{F8FFF8} Ours (Full)                     & 79.0          & 77.3          & 71.8          & 68.7          & 69.4          & 71.9           & 69.8          & 68.2          & 63.7          & 61.4          & 61.2          & 63.7           & 63.2          & 52.9          & 29.7          & 18.6          & 18.8          & 37.1           \\
			\hline
			\rowcolor[HTML]{F8FBFF} Naive Solution                  & 59.7          & 56.0          & 48.8          & 41.1          & 46.4          & 48.5           & 53.8          & 50.9          & 43.8          & 38.7          & 42.2          & 44.2           & 30.3          & 17.7          & 7.4           & 1.4           & 4.6           & 12.5           \\
			\rowcolor[HTML]{F8FBFF} \multicolumn{1}{c}{vs. Full}   & \textcolor{red}{-19.3} & \textcolor{red}{-21.3} & \textcolor{red}{-23.0} & \textcolor{red}{-27.6} & \textcolor{red}{-23.0} & \textcolor{red}{-23.4} & \textcolor{red}{-16.0} & \textcolor{red}{-17.3} & \textcolor{red}{-19.9} & \textcolor{red}{-22.7} & \textcolor{red}{-19.0} & \textcolor{red}{-19.5} & \textcolor{red}{-32.9} & \textcolor{red}{-35.2} & \textcolor{red}{-22.3} & \textcolor{red}{-17.2} & \textcolor{red}{-14.2} & \textcolor{red}{-24.6} \\

			\rowcolor[HTML]{FFFCF5} w/o Hard Text Attribute Binding             & 75.5          & 72.3          & 68.1          & 64.6          & 63.2          & 67.2           & 66.9          & 65.3          & 60.6          & 58.1          & 56.1          & 60.0           & 58.1          & 41.8          & 24.3          & 11.4          & 9.7           & 29.5           \\
			\rowcolor[HTML]{FFFCF5} \multicolumn{1}{c}{vs. Full}   & \textcolor{red}{-3.5} & \textcolor{red}{-5.0} & \textcolor{red}{-3.7} & \textcolor{red}{-4.1} & \textcolor{red}{-6.2} & \textcolor{red}{-4.7} & \textcolor{red}{-2.9} & \textcolor{red}{-2.9} & \textcolor{red}{-3.1} & \textcolor{red}{-3.3} & \textcolor{red}{-5.1} & \textcolor{red}{-3.7} & \textcolor{red}{-5.1} & \textcolor{red}{-11.1} & \textcolor{red}{-5.4} & \textcolor{red}{-7.2} & \textcolor{red}{-9.1} & \textcolor{red}{-7.6} \\

			\toprule[0.8pt]
		\end{tabular}
	}
	\vspace{-4.0mm}
\end{table*}

%% file: Tables/Ablation_Image_Binding.tex
\begin{table*}[t!]
	\centering
	\caption{\textbf{Ablation study on Image Attribute Binding (\S~\ref{sec:image_binding}).} $\mathcal{L}_i$ means that the count of instances needed to generate is \textbf{i}.}\label{tab:ablation_image_binding}
	\vspace{-3.5mm}
	\centering
	\resizebox{\textwidth}{!}{
		\setlength{\tabcolsep}{3.8pt}
		\footnotesize
		\renewcommand\arraystretch{1.1}
		\begin{tabular}{ccccccccccccccccccc}
			\bottomrule[1pt]\rowcolor[HTML]{FAFAFA}
			                             & \multicolumn{6}{c}{Instance Success Ratio (\%)$\uparrow$} & \multicolumn{6}{c}{Mean Intersection over Union (\%)$\uparrow$} & \multicolumn{6}{c}{Image Success Ratio (\%)$\uparrow$}                                                                                                                                                                                                                                                                               \\
			\cmidrule(lr){1-1} \cmidrule(lr){2-7} \cmidrule(lr){8-13} \cmidrule(lr){14-19}
			Method                       & $\mathcal{L}_2$                                           & $\mathcal{L}_3$                                                 & $\mathcal{L}_4$                                        & $\mathcal{L}_5$ & $\mathcal{L}_6$ & $\mathcal{AVG}$ & $\mathcal{L}_2$ & $\mathcal{L}_3$ & $\mathcal{L}_4$ & $\mathcal{L}_5$ & $\mathcal{L}_6$ & $\mathcal{AVG}$ & $\mathcal{L}_2$ & $\mathcal{L}_3$ & $\mathcal{L}_4$ & $\mathcal{L}_5$ & $\mathcal{L}_6$ & $\mathcal{AVG}$ \\ \toprule[0.8pt]

			w/o Hard Image Bind          & 71.3                                                      & 69.5                                                            & 56.9                                                   & 51.4            & 47.6            & 56.2            & 63.8            & 61.3            & 52.3            & 47.5            & 43.9            & 51.2            & 53.6            & 37.9            & 16.2            & 4.3             & 3.9             & 23.6            \\
			Hard Image Bind Input Layer  & 71.0                                                      & 68.4                                                            & 56.6                                                   & 52.9            & 50.3            & 57.1            & 62.4            & 60.5            & 50.4            & 47.2            & 45.4            & 50.9            & 51.0            & 34.0            & 14.9            & 5.7             & 5.2             & 22.5            \\
			\textbf{Hard Image Bind Middle Layer (Ours)} & \textbf{79.0}                                             & \textbf{77.3}                                                   & \textbf{71.8}                                          & \textbf{68.7}   & \textbf{69.4}   & \textbf{71.9}   & \textbf{69.8}   & \textbf{68.2}   & \textbf{63.7}   & \textbf{61.4}   & \textbf{61.2}   & \textbf{63.7}   & \textbf{63.2}   & \textbf{52.9}   & \textbf{29.7}   & \textbf{18.6}   & \textbf{18.8}   & \textbf{37.1}   \\
			Hard Image Bind Output Layer & 64.5                                                      & 63.8                                                            & 57.4                                                   & 49.4            & 58.3            & 57.6            & 56.9            & 57.2            & 51.8            & 45.0            & 52.1            & 51.7            & 43.9            & 28.8            & 14.9            & 2.9             & 7.8             & 20.0            \\

			\toprule[0.8pt]
		\end{tabular}
	}
	\vspace{-4.0mm}
\end{table*}

%% file: Figures/04_Ablation_Image_Binding.tex
\begin{figure*}[ht!]
	\includegraphics[width=1.0\linewidth]{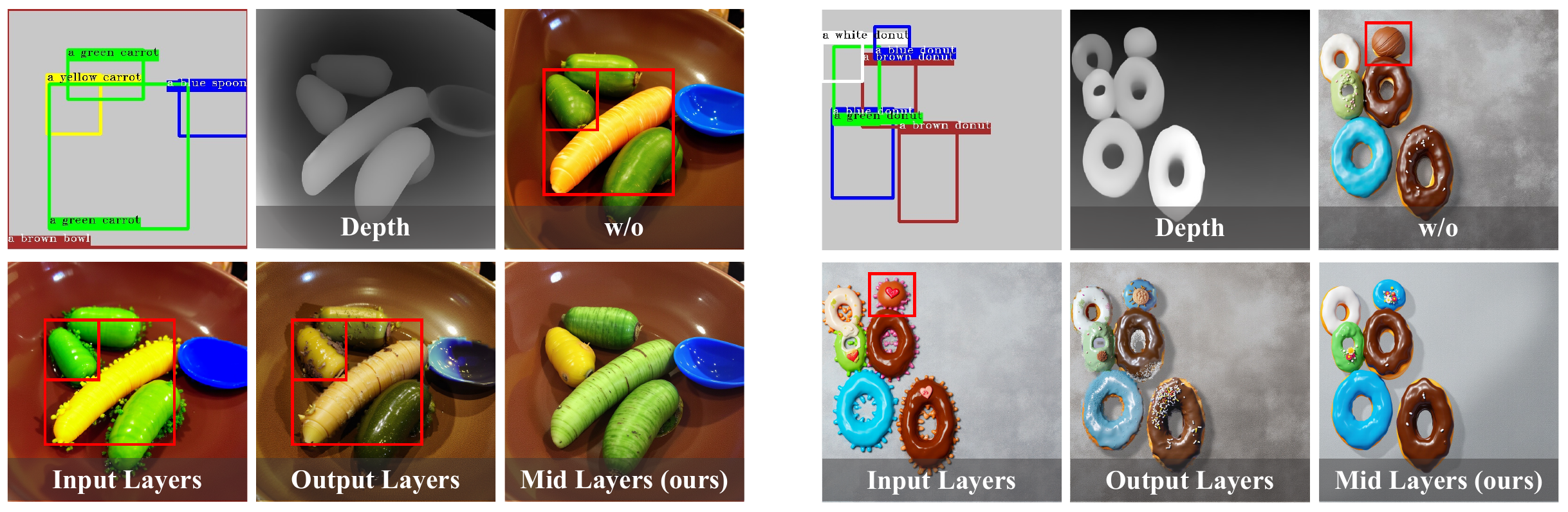}
	\vspace{-6mm}
	\caption{
		\textbf{Ablation study on Hard Image Attribute Binding (\S~\ref{sec:image_binding}).} 
	}
	\vspace{-4mm}
	\label{fig:ablation_image_binding}
\end{figure*}

%% file: Figures/04_Ablation_Text_Binding.tex
\begin{figure}[tb]
	\centering
	\includegraphics[width=\linewidth]{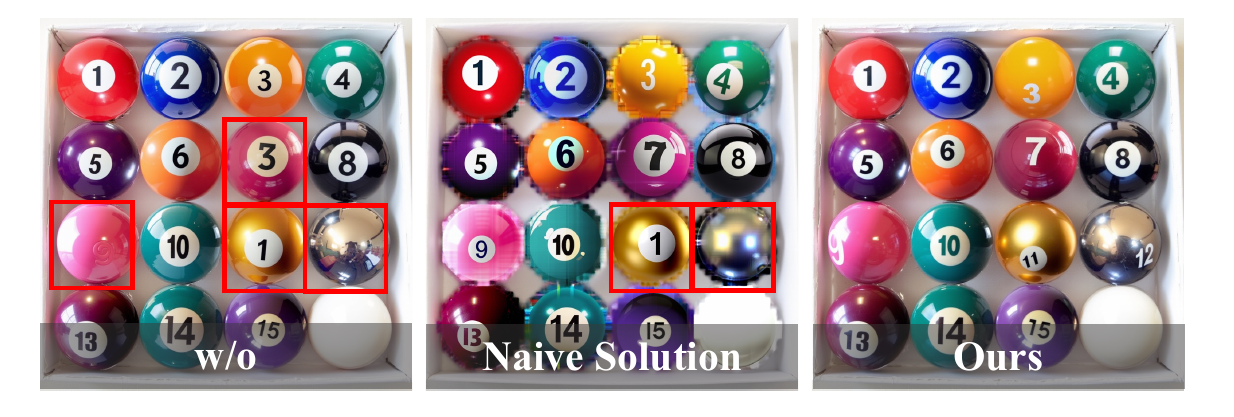}
	\vspace{-6mm}
	\caption{
		\textbf{Ablation study on Hard Text Attribute Binding (\S~\ref{sec:hard_text_binding}). 
}We use the same layout from Fig.~\ref{fig:teaser} for testing. Due to space limitations in the main text, additional image results are provided in the Supplementary Materials.
	}
	\vspace{-6mm}
	\label{fig:ablation_text_binding}
\end{figure}

%% file: Sections/5_conclusion.tex
\vspace{-0.5em}
\section{Conclusion}
\label{sec:conclusion}
We present DreamRenderer, a \textbf{plug-and-play} approach that enables depth- and canny-conditioned generation to control the content of specific regions and instances, without compromising the original model’s image quality. Our work makes two key contributions. First, we introduce a novel Hard Text Attribute Binding mechanism that employs Bridge Image Tokens, ensuring each instance’s text embedding binds the correct visual information during Joint Attention. Second, through an experimental analysis of FLUX’s individual layers, we apply hard image attribute binding only to the vital layers, maintaining precise instance-level control while preserving global image coherence. Extensive experiments on the COCO-POS and COCO-MIG benchmarks demonstrate DreamRenderer’s superior performance. In the depth-guided setting, our method achieves 62.50\% SR, 94.51\% ISR, and 84.36\% MIoU, substantially outperforming existing approaches. Even under the more challenging canny-guided setting, DreamRenderer remains robust, achieving 74.61\% ISR, and 66.95\% MIoU. Furthermore, DreamRenderer can serve as a re-renderer, significantly improving the accuracy of layout-to-image approaches. Its \textbf{training-free} nature allows DreamRenderer to be easily applied to various foundation models, offering high flexibility. In the future, we will further explore DreamRenderer’s integration with additional image-conditioned generation approaches.

%% file: Sections/6_appendix.tex

\newpage
\appendix
\renewcommand{\thepart}{} 
\renewcommand{\partname}{} 
\part{Appendix} 

\setcounter{figure}{0}
\setcounter{table}{0}
\renewcommand{\thetable}{\Alph{table}}
\renewcommand{\thefigure}{\Alph{figure}}

\section{More Qualitative Results}
\label{sec:sup_additional_qualitative_results}
\cref{fig:sub_generated_person,fig:sub_generated_Chinese_architecture,fig:sub_generated_design} presents additional qualitative results of our method. 
\cref{fig:sub_generated_design} showcases our method's effectiveness and flexibility in graphic design applications. The example demonstrates how DreamRenderer can generate distinct design variations with only little input modifications. In the top design, we generate an orange bold "Dream Big!" text in the center, and with just a small modification to the text prompt from ``Dream Big!'' to two different color text prompts (orange ``Dream'' and blue ``Renderer''), our method generates an entirely new variant (bottom).

As shown in \cref{fig:sub_generated_person}, we tackle the challenging task of simultaneously generating multiple specified person. Generating seven distinct person while maintaining consistent identity and appearance is particularly difficult, as it requires the model to understand and preserve individual characteristics across different poses and viewpoints. Our method successfully generates natural-looking results where not only is the identity consistently maintained across all instances, but the generated images also precisely align with the provided depth conditions. This demonstrates our model's robust capability in handling complex multi-instance person generation tasks.

\section{More Results on COCO-POS Benchmark}
\cref{fig:sub_coco_pos} presents the qualitative results of our method compared with FLUX~\cite{flux} and 3DIS~\cite{zhou20243dis} on both depth-guided and canny-guided generation. As shown in the figure, our DreamRenderer consistently outperforms both FLUX and 3DIS, particularly when generating multiple instances. In the \textbf{depth-guided} scenarios (top rows), our method accurately preserves the spatial relationships indicated in the depth maps while ensuring that each instance's attributes are correctly rendered according to the textual descriptions. The baseline methods struggle with attribute entanglement, often generating instances with incorrect colors, patterns, or other visual characteristics.

For the more challenging \textbf{canny-guided} generation (bottom rows), the performance gap is even more pronounced. While FLUX and 3DIS frequently produce instances with misaligned attributes or distorted appearances, our method maintains attribute fidelity even with minimal structural guidance from canny edges. This demonstrates the effectiveness of our \textbf{Hard Text Attribute Binding} mechanism, which ensures each instance's text embedding correctly binds with its corresponding visual features during the generation process.

Notably, our method achieves these improvements without compromising image quality. The generated images exhibit clear details, natural textures, and coherent global compositions, demonstrating that our \textbf{Image Attribute Binding} approach successfully preserves the model's inherent rendering capabilities while enhancing attribute control.

\section{More Results on COCO-MIG Benchmark}
\cref{fig:more_gligen}, \cref{fig:more_instancediff}, \cref{fig:more_migc}, and \cref{fig:more_3dis} present comparative results obtained by applying DreamRenderer to re-render outputs from various layout-to-image methods. The red boxes indicate that all methods exhibit challenges in attribute binding to varying degrees.

\textbf{GLIGEN}~\cite{gligen}, one of the earliest methods with layout control, shows the most severe attribute confusion, often generating objects with incorrect colors or patterns and struggling with spatial consistency. \textbf{InstanceDiffusion}~\cite{instancediffusion} improves instance separation but still struggles with attribute binding across multiple instances, and notably suffers from lower visual quality with blurry textures and less detailed renderings. \textbf{MIGC}~\cite{migc} produces high-resolution results but frequently fails to adhere to depth conditions properly, and often generates images with overly saturated colors and unrealistic brightness. Even the most advanced method \textbf{3DIS}~\cite{zhou20243dis} exhibits significant attribute binding errors when handling multiple instances with similar categories but different properties.

Our DreamRenderer consistently enhances performance across all methods by ensuring accurate attribute binding while preserving image quality. The improvements become more pronounced when controlling multiple similar instances with different visual properties, confirming our method's effectiveness in addressing attribute entanglement regardless of the underlying architecture.

\begin{figure*}[h]
	\centering
	\includegraphics[width=\textwidth]{Sources/supp/self_generated_design_renderer.pdf}
	\caption{\textbf{Qualitative results of our method on multi-instance design generation. (\S~\ref{sec:sup_additional_qualitative_results})} Our model enables efficient design iterations with precise control, enabling designers to explore multiple design layout by only little input modifications.}
	\label{fig:sub_generated_design}
\end{figure*}
\begin{figure*}[h]
	\centering
	\includegraphics[width=\textwidth]{Sources/supp/self_generated_person.pdf}
	\caption{\textbf{Qualitative results of our method on multi-instance person generation. (\S~\ref{sec:sup_additional_qualitative_results})} Our model successfully generates 7 different persons simultaneously, which is a notably \textbf{challenging} task in image generation. The results demonstrate consistent identity preservation across all instances while maintaining natural appearance variations. Each generated image accurately follows the corresponding depth condition, showing our model's ability to handle complex spatial relationships and viewpoint variations.}
	\label{fig:sub_generated_person}
\end{figure*}
\begin{figure*}[h]
	\centering
	\includegraphics[width=\textwidth]{Sources/supp/self_generated_building.pdf}
	\caption{\textbf{Qualitative results of our method on different architecture style generation. (\S~\ref{sec:sup_additional_qualitative_results})} Our model successfully generates images with the same layout input, demonstrating its versatility in capturing and reproducing diverse artistic styles.}
	\label{fig:sub_generated_Chinese_architecture}
    \vspace{-5pt}
\end{figure*}

\begin{figure*}[h]
	\centering
	\includegraphics[width=\textwidth]{Sources/supp/coco_pos.pdf}
	\caption{\textbf{Qualitative comparison with FLUX and 3DIS on depth-guided (top) and canny-guided (bottom) generation. (\S~\ref{sec:sup_additional_qualitative_results})} Our method produces images with more accurate attributes and better visual quality, while baseline methods often exhibit color and pattern inconsistencies with text prompts.
	}
	\label{fig:sub_coco_pos}
\end{figure*}

\input{Figures/supp/MORE_GLIGEN}
\input{Figures/supp/More_InstanceDiff}
\input{Figures/supp/MORE_MIGC}
\input{Figures/supp/MORE_3DIS}

\section{More Results on Hard Text Attribute Binding}
\label{sec:sup_ablation_hard_text_binding}

\cref{fig:sub_ablation_text_binding} demonstrates the effectiveness of our Hard Text Attribute Binding mechanism. \textbf{The naive approach} generally preserves basic attributes but suffers from severely degraded image quality with noticeable artifacts. The model \textbf{without} Hard Text Attribute Binding produces visually appealing images but frequently fails to correctly bind text attributes to the generated content, resulting in misaligned visual elements. In contrast, our full model with \textbf{Hard Text Attribute Binding} achieves both high image quality and accurate attribute preservation. Comparing the three approaches side by side, we observe that our method successfully addresses the limitations of both alternative approaches, delivering consistent text-image alignment without compromising visual fidelity.

\begin{figure*}[h]
	\centering
    \vspace{-5pt}
	\includegraphics[width=\textwidth]{Sources/supp/ablation_hard_text.pdf}
    \vspace{-5pt}
	\caption{\textbf{Ablation study on the Hard Text Attribute Binding mechanism. (\S~\ref{sec:sup_ablation_hard_text_binding})} Top: Results from the naive approach, which maintains basic attribute correctness but produces poor image quality with significant artifacts. Bottom: Results without hard text binding, showing good visual quality but frequent attribute binding failures. Results from our full model, demonstrating both high-quality image generation and accurate text attribute binding. The comparison highlights how our method effectively balances visual quality and text-prompt adherence.}
	\label{fig:sub_ablation_text_binding}
\end{figure*}

\section{Limitations}
Despite the significant advancements achieved by DreamRenderer in multi-instance generation control, several limitations persist. For canny-guided generation, our method's performance is less robust compared to depth-guided generation, primarily constrained by the capabilities of the underlying FLUX-Canny model, as evidenced by the results in body part's Tab. 1. Furthermore, we observe a substantial decrease in the success ratio as the number of controlled instances increases, a phenomenon particularly pronounced in canny-guided generation, where the success rate drops from 23.28\% with two instances to considerably lower values with additional instances. Although the Hard Text Attribute Binding mechanism significantly improves attribute binding accuracy, the attribute entanglement issue remains not fully resolved when handling complex scenes with multiple overlapping instances, indicating room for further improvement in this domain.

%% file: Figures/supp/MORE_GLIGEN.tex
\begin{figure*}[ht!]
	\centering
	\includegraphics[width=1.0\linewidth]{Sources/supp/more_GLIGEN.pdf}
    \vspace{-5mm}
	\caption{\textbf{Additional qualitative comparison on the COCO-MIG benchmark. (\S~\ref{sec:sup_additional_qualitative_results})} We show more results of re-rendering on GLIGEN~\cite{gligen}. We highlight with red boxes the areas where the compared method exhibits noticeable attribute generation errors.
	}
	\label{fig:more_gligen}
\end{figure*}

%% file: Figures/supp/More_InstanceDiff.tex
\begin{figure*}[ht!]
	\centering
	\includegraphics[width=1.0\linewidth]{Sources/supp/more_InstanceDiff.pdf}
    \vspace{-5mm}
	\caption{\textbf{Additional qualitative comparison on the COCO-MIG benchmark. (\S~\ref{sec:sup_additional_qualitative_results})} We show more results of re-rendering on InstanceDiffusion~\cite{instancediffusion}. We highlight with red boxes the areas where the compared method exhibits noticeable attribute generation errors.
	}
	\label{fig:more_instancediff}
\end{figure*}

%% file: Figures/supp/MORE_MIGC.tex
\begin{figure*}[ht!]
	\centering
	\includegraphics[width=1.0\linewidth]{Sources/supp/more_MIGC.pdf}
    \vspace{-5mm}
	\caption{\textbf{Additional qualitative comparison on the COCO-MIG benchmark. (\S~\ref{sec:sup_additional_qualitative_results})} We show more results of re-rendering on MIGC~\cite{migc}. We highlight with red boxes the areas where the compared method exhibits noticeable attribute generation errors.
	}
	\label{fig:more_migc}
\end{figure*}

%% file: Figures/supp/MORE_3DIS.tex
\begin{figure*}[h]
	\centering
	\includegraphics[width=1.0\linewidth]{Sources/supp/more_3DIS.pdf}
    \vspace{-5mm}
	\caption{\textbf{Additional qualitative comparison on the COCO-MIG benchmark. (\S~\ref{sec:sup_additional_qualitative_results})} We show more results of re-rendering on 3DIS~\cite{zhou20243dis}. We highlight with red boxes the areas where the compared method exhibits noticeable attribute generation errors.
	}
	\label{fig:more_3dis}
    \vspace{-5pt}
\end{figure*}